\documentclass{ecai} 

\usepackage{latexsym}
\usepackage{amssymb}
\usepackage{amsmath}
\usepackage{amsthm}
\usepackage{booktabs}
\usepackage{enumitem}
\usepackage{graphicx}
\usepackage{color}
\usepackage{bm}

\newcommand{\BibTeX}{B\kern-.05em{\sc i\kern-.025em b}\kern-.08em\TeX}

\begin{document}
\pretolerance=10000

\begin{frontmatter}

\paperid{12} 


\title{
Towards Interpretable Concept Learning over Time Series via Temporal Logic Semantics}


\author[A]{\fnms{Irene}~\snm{Ferfoglia}\orcid{0000-0003-1585-6576}\thanks{Corresponding Author. Email: irene.ferfoglia@phd.units.it}} 
\author[A]{\fnms{Simone}~\snm{Silvetti}\orcid{0000-0001-8048-9317}}
\author[A]{\fnms{Gaia}~\snm{Saveri}\orcid{0009-0003-2948-7705}} 
\author[A]{\fnms{Laura}~\snm{Nenzi}\orcid{0000-0003-2263-9342}} 
\author[A]{\fnms{Luca}~\snm{Bortolussi}\orcid{0000-0001-8874-4001}} 

\address[A]{Università degli Studi di Trieste}


\begin{abstract}
Time series classification is a task of paramount importance, as this kind of data often arises in safety-critical applications. However, it is typically tackled with black-box deep learning methods, making it hard for humans to understand the rationale behind their output. 
To take on this challenge, we propose a neuro-symbolic framework that unifies classification and explanation through direct embedding of trajectories into a space of Signal Temporal Logic (STL) concepts. By introducing a novel STL-inspired kernel that maps raw time series to their alignment with predefined STL formulae, our model jointly optimises for accuracy and interpretability, as each prediction is accompanied by the most relevant logical concepts that characterise it.
This enables classification grounded in human-interpretable temporal patterns and produces both local and global symbolic explanations. Early results show competitive performance while offering high-quality logical justifications for model decisions.
\end{abstract}
\end{frontmatter}


\section{Introduction} \label{sec:intro}

Time series data are pervasive across domains, from industrial monitoring to safety-critical applications. While machine learning models have demonstrated strong performance in analysing such data \cite{tsc-survey,dl-tsc}, their black-box nature poses significant challenges to transparency and trust, especially in high-stakes contexts governed by ethical \cite{ethics1, ethics2} and legal \cite{legal, gdpr} concerns.

Explainable AI (XAI) techniques for time series often adapt methods originally developed for images or text \cite{tsc-xai-survey, rojat2021explainable}. Prominent approaches include attribution-based methods \cite{Karim_2018, hsieh2020explainable}, which highlight influential subsequences in the input. However, these post-hoc techniques have been criticised for their potential to misrepresent the model’s internal reasoning \cite{tsc-posthoc-survey}, and their explanations can be difficult to interpret due to the abstract nature of time series patterns.

Symbolic formalisms such as Signal Temporal Logic (STL) \cite{temporal-logic, stl} offer a compelling alternative for describing temporal behaviours in an interpretable and formal way. STL has been widely used in requirement mining and rule extraction \cite{roge, ir-bo, bombara, enumerate-stl}, but most prior work provides global descriptions of class behaviour, without explaining individual predictions. More recent neuro-symbolic models attempt to embed STL in decision tree structures or neural architectures \cite{nesy-tsc-one, nesy-tsc-threev2}, yet they still lack the flexibility to generate instance-level explanations aligned with input semantics.

We address this gap through a concept-based approach \cite{ghorbani2019automatic, NEURIPS2020_ecb287ff} that integrates STL formulae directly into the model as interpretable units. Unlike prior works that generate STL-based classifiers or tree-based rules \cite{nesy-tsc-one, nesy-tsc-two}, our method computes both local and global explanations by embedding time series and STL concepts into a shared space using a robustness-inspired kernel \cite{stl-kernel}. This enables predictions to be interpreted directly in terms of temporal logic conditions, with relevance scores computed via a cross-attention mechanism. Importantly, our method is designed to handle both multivariate signals, by supporting STL formulae over multiple variables, and multiclass settings, through class-specific discriminability and attention mechanisms.



Our intended contributions are: (i) a novel time series classifier leveraging STL-based kernels for interpretable concept learning; (ii) local and global explanations via STL formulae, clarifying individual predictions and overall model behaviour; and (iii) empirical validation on real-world cyber-physical systems and scalability benchmarks.

\section{Background} \label{sec:background}
\paragraph{Concept-based models} 
aim to make machine learning inherently interpretable by incorporating human-understandable concepts directly into the learning process \cite{ghorbani2019automatic, NEURIPS2020_ecb287ff}. Unlike post-hoc explanation methods, these models use concepts, often representing domain knowledge, as part of the prediction pipeline. 

\paragraph{Signal Temporal Logic (STL)} is a linear-time temporal logic for specifying properties of continuous-time trajectories $\tau: I \rightarrow \mathcal{X}$ \cite{stl}. Its syntax combines Boolean and temporal operators:
\begin{equation*}
\varphi := \top \mid \pi \mid \lnot\varphi \mid \varphi_1\land\varphi_2 \mid \varphi_1\bm{U}_{[a,b]}\varphi_2
\end{equation*}
where predicates $\pi$ are inequalities over signal values. Common derived operators include $\bm{F}_{[a,b]}$ (eventually), $\bm{G}_{[a,b]}$ (globally), and $\bm{U}_{[a,b]}$ (until). 
STL supports both Boolean satisfaction and quantitative semantics via robustness $\rho(\varphi,\tau,t)$, which measures how strongly a formula holds.

\section{Trajectory embedding kernel}
A kernel for Signal Temporal Logic (STL) can be constructed using the quantitative robustness semantics, where each formula $\varphi$ is viewed as a functional $\rho(\varphi, \cdot): \mathcal{T} \rightarrow \mathbb{R}$ mapping trajectories to real values. By equipping the trajectory space $\mathcal{T}$ with a probability measure $\mu_0$, a similarity kernel between two formulae $\varphi$ and $\psi$ is defined as:
\small
\begin{equation*}
k(\varphi, \psi) = \langle \rho(\varphi, \cdot), \rho(\psi, \cdot) \rangle = \int_{\tau\in \mathcal{T}} \rho(\varphi, \tau) \rho(\psi, \tau) \, d\mu_0(\tau)
\end{equation*}
This kernel captures behavioural similarity based on how formulae evaluate over likely trajectories, enabling a continuous embedding of symbolic logic into a Hilbert space. Extending this idea, we define a dual embedding for trajectories by introducing a functional $\rho_\tau : \mathcal{T} \rightarrow [-1, 1]$ given by:
\begin{equation*} \label{eq:rho_tau}
\rho_{\tau}(\xi) = 2\cdot \exp \bigg( - \frac{d_{\tau}(\xi)}{\varepsilon}\bigg)-1 
\end{equation*}
where $d_\tau(\xi) = \| \xi - \tau \|^2$ and $\varepsilon > 0$ controls locality. This yields a cross-kernel between a trajectory $\tau$ and a formula $\varphi$:
\begin{equation*}
k(\tau, \varphi) = \int_{\xi \in \mathcal{T}} \rho_\tau(\xi) \rho(\varphi, \xi) \, d\mu_0(\xi)
\label{eq:traj-formula-kernel}
\end{equation*}
allowing both formulae and trajectories to be embedded in a shared semantic space. This facilitates alignment between data and logic, forming the mathematical foundation for our interpretable concept-based classification framework.

\section{The model}
In our framework, concepts are defined as human-interpretable temporal patterns encoded as STL formulae. To construct a semantically rich yet interpretable concept set, we sample candidate formulae from a probabilistic grammar with constraints on the number of variables and syntactic complexity. Each formula is evaluated on a fixed dataset of time series signals using robustness semantics, resulting in robustness signature vectors that capture their behaviour across trajectories. Concept selection proceeds iteratively: new candidates are retained only if their signatures are sufficiently dissimilar (measured via cosine similarity) from those already selected. Simpler formulae are preferred in the presence of redundancy. Each formula may involve one or more signal dimensions, allowing the model to express and attend to complex relationships in multivariate inputs. The final set is generated once using a fixed time window, and can be rescaled for datasets of different lengths by adjusting temporal thresholds. This ensures that the selected formulae are both diverse and interpretable, capturing a wide range of temporal behaviours.

\subsection{Architecture}
Our architecture performs interpretable time series classification by integrating symbolic reasoning based on Signal Temporal Logic (STL) with neural representations of trajectories. It operates over a fixed set of human-interpretable STL formulae $\Phi_C = \{ \varphi_1, \dots, \varphi_n \}$, used as temporal logic concepts. Given an input time series $\bm{x}$, we compute its embedding $\mathcal{H}_{\bm{x}} \in \mathbb{R}^n$ by evaluating the robustness $\rho(\varphi_i, \bm{x})$ for each concept $\varphi_i \in \Phi_C$. This results in a fully interpretable vector where each dimension reflects the degree to which a known temporal property is satisfied. To assess how informative each concept is for classification, we compute class-specific discriminability scores. The model supports an arbitrary number of classes $K$, enabling it to scale naturally to multiclass classification tasks. For each class $k$ and formula $\varphi_i$, we define a discriminability matrix $\mathcal{G}_{\bm{x}} \in \mathbb{R}^{n \times K}$ with entries
\begin{equation*}
 \mathcal{G}_i^{k} = \frac{\rho(\varphi_i, \bm{x}) - \mu_{k^*, i}}{\sigma_{k^*, i} + \varepsilon_G}
 \end{equation*}
where $k^* \neq k$, and $\mu_{k^*, i}, \sigma_{k^*, i}$ are the mean and standard deviation of robustness values over class $k^*$, and $\varepsilon_G$ is a small constant. This normalisation highlights whether a concept is atypical for other classes. A soft attention mechanism identifies the most relevant concepts via:
\begin{equation*}
\alpha(\bm{x}) = \text{softmax} \left( \mathcal{H}_{\bm{x}} / T_{attn} \right) \in \mathbb{R}^n
\end{equation*}
where $T_{attn}$ is a temperature parameter. The weighted combination of attention and discriminability produces the modulated score matrix:
\begin{equation*} \label{eq:z}
\bm{z}(\bm{x}) = \alpha(\bm{x}) \odot \mathcal{G}_{\bm{x}} \in \mathbb{R}^{n \times K}   
\end{equation*}
which is flattened and passed through a linear layer to obtain the class prediction.
This architecture ensures interpretability both through attention weights, indicating which concepts influenced the prediction, and through discriminability scores, explaining why those concepts support or oppose the predicted class.

\subsection{Explanation extraction}
Explanations are derived by combining three elements: attention weights $\alpha(\bm{x})$, discriminability scores $\mathcal{G}_{\bm{x}}$, and classifier weights $\bm{W}$. For a predicted class $\hat{y}$, the explanation scores are computed as $\bm{r} = \bm{W}{\hat{y}} \cdot \bm{z}(\bm{x}){\hat{y}}$, where $\bm{z}(\bm{x})$ reflects the relevance of each concept. The most influential STL formulae are selected either by choosing the top-$\gamma$ highest absolute relevance scores or by applying a cumulative relevance threshold. These selected formulae $\mathcal{F}_{\bm{x}}$ are then simplified by adjusting thresholds, polarities, and structure, resulting in a refined set $\mathcal{F}’_{\bm{x}}$. The local explanation is a conjunction:
\begin{equation*}
E_l(\bm{x}, \hat y) = \bigwedge_{\varphi_i' \in \mathcal{F}'_{\bm{x}}} \varphi_i'
\end{equation*}
which serves as a sufficient, human-readable condition for predicting $\hat{y}$. To extract global explanations for a class $k$, all local explanations of training samples with label $k$ are aggregated into a set $\mathcal{F}_k$. A discriminative subset $\mathcal{F}’_k$ is then selected by solving a minimum-cost set cover problem, where cost reflects syntactic complexity. The resulting global explanation is a disjunction:
\begin{equation*}
E_g(k) = \bigvee_{\varphi_i \in \mathcal{F}'_k} \varphi_{i}
\end{equation*}
capturing sufficient symbolic conditions for class membership. \\To further enhance clarity and relevance, all formulae undergo three post-processing steps: (i) logical simplification (e.g. removing redundancies), (ii) threshold adjustment and polarity correction to better align with data, and (iii) data-aware pruning of consistently true/false subformulae. These steps improve the interpretability and discriminative power of both local and global explanations.

\section{Planned experimental evaluation}
To evaluate both the predictive performance and interpretability of our approach, we plan to conduct experiments on a wide selection of datasets from the UCR Time Series Archive \cite{UCR}. This benchmark includes over $140$ univariate and multivariate classification tasks spanning diverse domains such as physiological monitoring, sensor data, and motion capture, and offers a robust testbed for assessing generalisability.

Our method will be compared against a broad set of state-of-the-art time series classification models, including both high-performing black-box architectures and models explicitly designed for interpretability. This comparison will enable us to assess trade-offs between accuracy and explainability across modelling paradigms. All models will be trained and evaluated using the standard train/test splits provided by the UCR repository, ensuring consistency with prior benchmarks \cite{tsc-survey, Ruiz_Flynn_Large_Middlehurst_Bagnall_2021}. 

\section{Conclusions}
We have introduced a neuro-symbolic framework for time series classification that leverages Signal Temporal Logic (STL) as a foundation for interpretable concept learning. By embedding both trajectories and STL formulae into a shared semantic space, our approach enables faithful, logic-grounded explanations at both local and global levels. Each prediction is accompanied by temporally meaningful logical conditions, supporting model transparency and potential downstream use in high-stakes or regulated settings.

 



\begin{ack}
This study was carried out within the PNRR research activities of the consortium iNEST (Interconnected North-Est Innovation Ecosystem) funded by the European Union Next-GenerationEU (Piano Nazionale di Ripresa e Resilienza (PNRR) – Missione 4 Componente 2, Investimento 1.5 – D.D. 1058 23/06/2022, ECS\_00000043). This manuscript reflects only the Authors’ views and opinions, neither the European Union nor the European Commission can be considered responsible for them.
\end{ack}



\bibliography{ecai-template/paper}

\end{document}